\documentclass[letterpaper, 10 pt, conference]{ieeeconf}  

\usepackage{graphics} 
\usepackage{epsfig} 
\usepackage{mathptmx} 
\usepackage{times} 
\usepackage{amsmath} 
\usepackage{amssymb}  
\usepackage{ulem}
\usepackage[colorlinks=true, bookmarks=true]{hyperref}
\usepackage{cite}
\usepackage{multirow}
\usepackage{comment}
\usepackage{multirow}
\usepackage{amsmath,amssymb,amsfonts}
\usepackage{algorithmic}
\usepackage{graphicx}
\usepackage{textcomp}
\usepackage{xcolor}
\usepackage{booktabs}
\usepackage{booktabs, multirow, makecell, threeparttable}
\usepackage{soul}
\usepackage{fontawesome}
\usepackage{ulem} 
\usepackage{cleveref}
\usepackage{svg}

\crefformat{figure}{Fig.~#2#1#3}  
\crefformat{section}{Sec.~#2#1#3}

\title{\LARGE \bf
XGrid-Mapping: Explicit–Implicit Hybrid Grid Submaps for Efficient Incremental Neural LiDAR Mapping
}

\author{Zeqing Song, Zhongmiao Yan, Junyuan Deng, Songpengcheng Xia, Xiang Mu, Jingyi Xu, Qi Wu, Ling Pei$^{*}$}

\begin{document}

\maketitle
 \thispagestyle{empty}
\pagestyle{empty}

\begin{abstract}

Large-scale incremental mapping is fundamental to the development of robust and reliable autonomous systems, as it underpins incremental environmental understanding with sequential inputs for navigation and decision-making. LiDAR is widely used for this purpose due to its accuracy and robustness. Recently, neural LiDAR mapping has shown impressive performance; however, most approaches rely on dense implicit representations and underutilize geometric structure, while existing voxel-guided methods struggle to achieve real-time performance. To address these challenges, we propose XGrid-Mapping, a hybrid grid framework that jointly exploits explicit and implicit representations for efficient neural LiDAR mapping. Specifically, the strategy combines a sparse grid, providing geometric priors and structural guidance, with an implicit dense grid that enriches scene representation. By coupling the VDB structure with a submap-based organization, the framework reduces computational load and enables efficient incremental mapping on a large scale. To mitigate discontinuities across submaps, we introduce a distillation-based overlap alignment strategy, in which preceding submaps supervise subsequent ones to ensure consistency in overlapping regions. To further enhance robustness and sampling efficiency, we incorporate a dynamic removal module. Extensive experiments show that our approach delivers superior mapping quality while overcoming the efficiency limitations of voxel-guided methods, thereby outperforming existing state-of-the-art mapping methods. \footnote{Code will be released upon paper acceptance.}

\end{abstract}

\section{INTRODUCTION}
\label{sec:intro}

The growing demand for autonomous driving and mobile robotics has highlighted the importance of efficient and accurate incremental large-scale outdoor LiDAR mapping. Reliable environment perception is essential to support downstream tasks such as simultaneous localization and mapping (SLAM), decision-making and path planning. 

Traditional point cloud fusion methods~\cite{VDBfusion, Voxblox, nvblox}, built on efficient spatial data structures~\cite{OctoMap, VOXELHASH, VDB}, achieve accurate LiDAR mapping but struggle in areas with sparse sensor coverage, resulting in incomplete maps. Recent work~\cite{deepsdf, NeRF, instant-ngp, DVGO, mipnerf360, pointnerf} introduced dense neural implicit representations for 3D geometry to obtain a richer and more expressive representation. Extended to large-scale outdoor environments with LiDAR inputs, methods~\cite{shine, pinslam, ucpinslam,clidslam, 3dslnr} adopt dense neural fields to obtain a continuous truncated signed distance field (TSDF). However, the dense sampling process usually lacks guidance from the existing map structure, which results in extensive sampling in free space and consequently leads to higher training costs, thereby failing to balance mapping quality and efficiency. In parallel, some work has explored the sparse implicit representation built on explicit voxel-guided sampling~\cite{plenoxels, voxfusion, h2mapping}. Subsequently, methods~\cite{nerfloam,s2kan} applied sparse implicit fields to outdoor LiDAR scenarios, achieving higher mapping quality with less training cost. However, their raycasting computational efficiency has gradually decreased as the scene increases, making real-time update impossible, limiting further exploration. In general, traditional point cloud fusion methods often yield incomplete maps, dense implicit approaches lack structural guidance and thus converge slowly. Sparse implicit representations provide structural priors for faster convergence but remain inefficient due to costly raycasting. 
\begin{figure}[t]
    \hspace{0.1cm} 
    \includegraphics[width=0.97\linewidth]{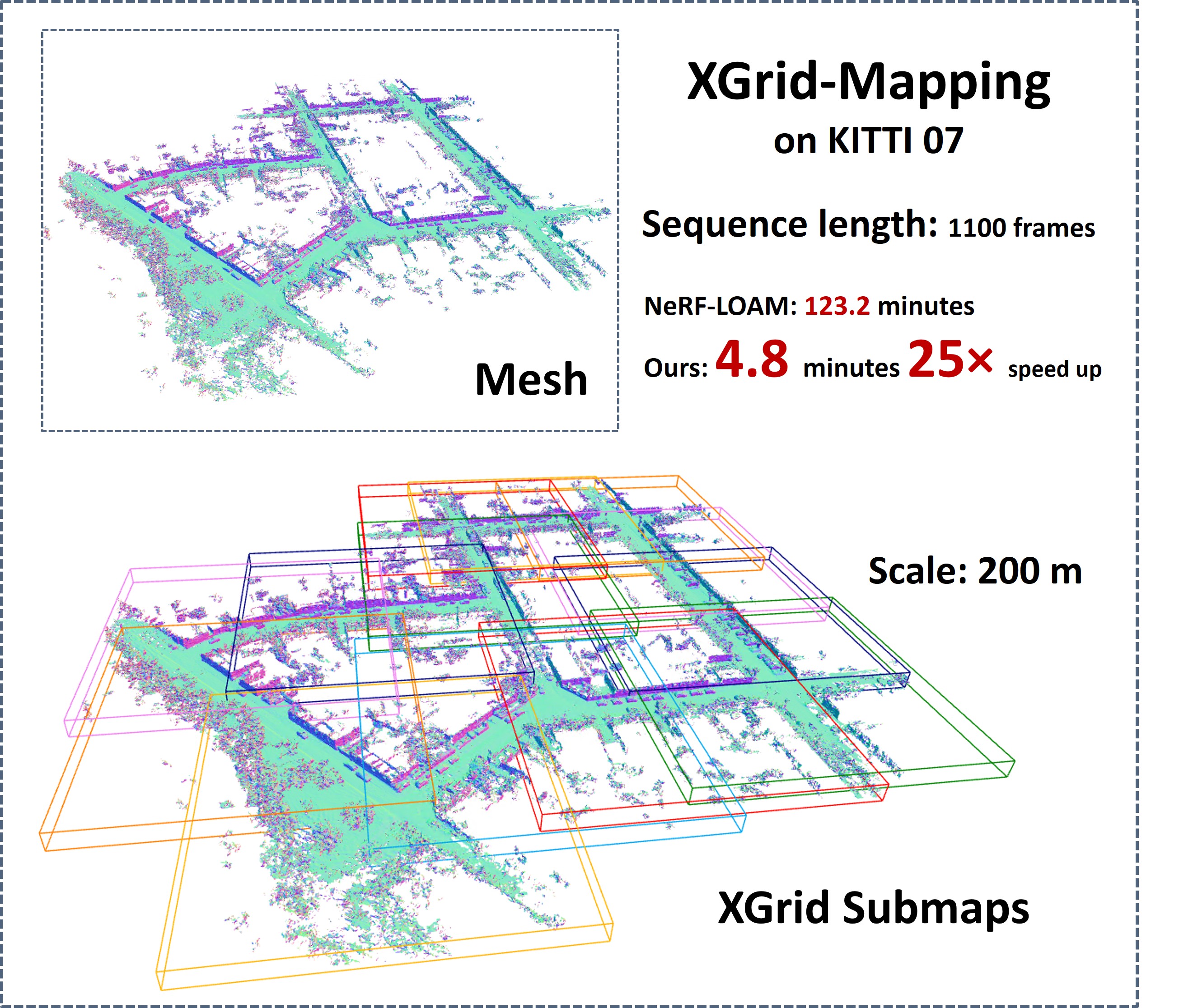}
     \caption{\textbf {Our mapping result and time efficiency on KITTI 07.} Boxes in different colors indicate different submaps created by our framework. Compared with NeRF-LOAM, our method achieves over 25× speed-up and reaches real-time performance.}
    \label{fig:intro}
    \vspace{-10pt}
\end{figure}



\begin{figure}[t]
 \centering
    \vspace{0.2cm}
    \includegraphics[width=1.0\linewidth]{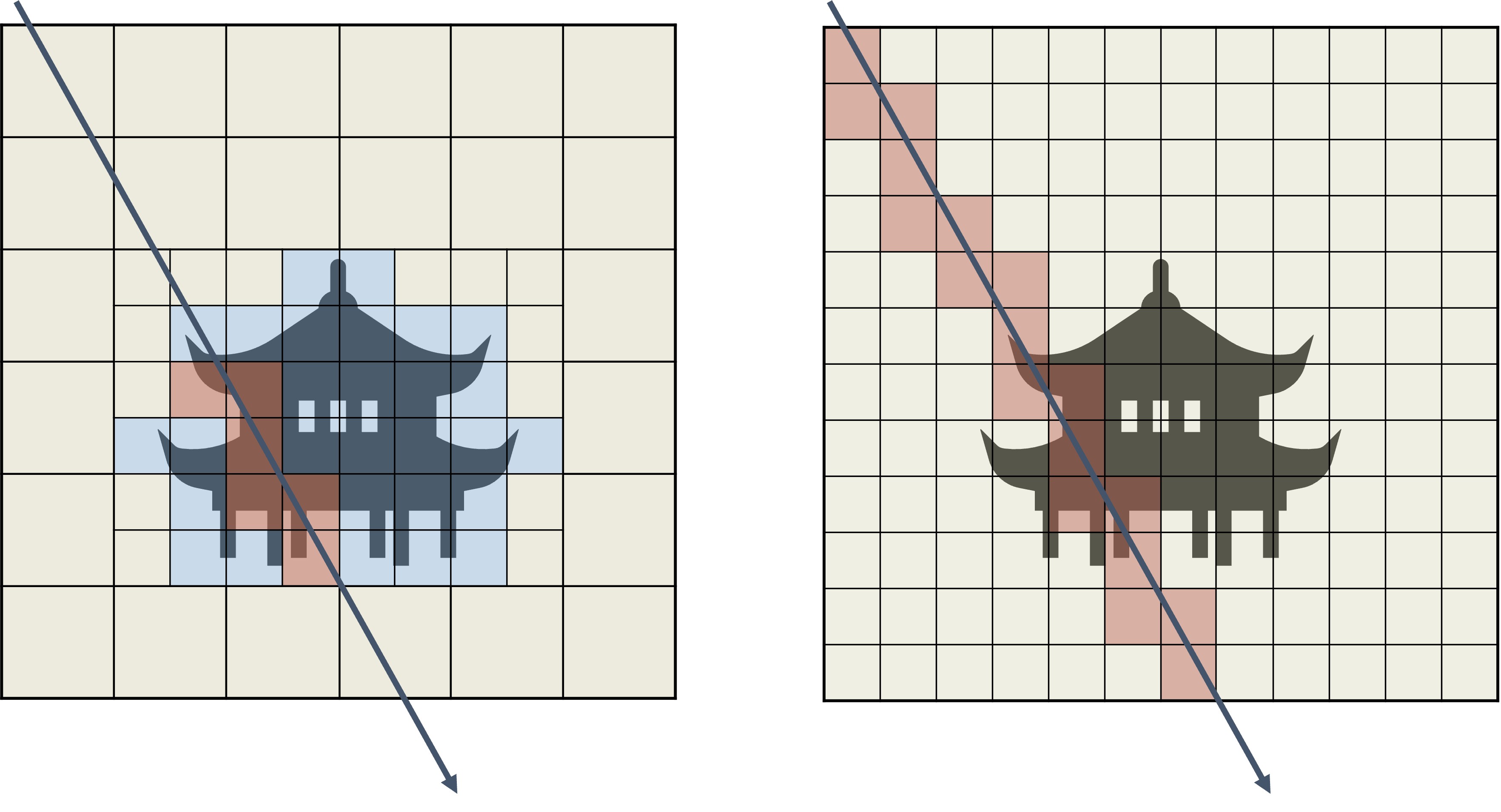}
    \vspace{-20pt}
     \caption{\textbf{Comparison of voxel-guided and dense sampling.} Brown cells denote sampled voxels, blue cells denote occupancy, and light cells denote unknown free space. Left: voxel-guided sampling with structural priors, ray traversal is guided by known occupancy, thus samples are concentrated near surfaces. Right: dense sampling without priors, where queries are spread along the ray for free space learning.
     }     
    \label{fig:related_work-voxel_intersect}
    \vspace{-10pt}
\end{figure}

In this paper, we propose XGrid-Mapping, an efficient mapping framework built on submaps of hybrid grid, with representative results shown in \cref{fig:intro}. The hybrid grid integrates an explicit sparse grid, which provides voxel-level structural guidance, with an implicit dense grid, which encodes fine-grained geometry to enrich representation accuracy. While such structural guidance enables more effective sampling, as shown in \cref{fig:related_work-voxel_intersect}, ray traversal becomes time-consuming as the scene grows more complex. To alleviate this issue, we introduce a submap formulation, which confines the sparse grid to localized regions and thereby reduces traversal overhead. To further enhance consistency, a distillation-based overlap alignment module enforces coherence across neighboring submaps, while a dynamic removal module improves robustness in complex environments. 






In summary, the primary contributions of our work are as follows:
\begin{itemize}
    \item We overcome the computational bottleneck of sparse implicit mapping by introducing a submap-based framework with an efficient sparse grid implementation, which enables updates in real time.  
    \item We design a hybrid grid training strategy that combines explicit structural guidance with implicit expressive ability. We further implement a dynamic removal module to improve mapping robustness and performance.  
    \item We introduce a distillation-based overlap alignment strategy that enforces feature consistency across submaps, ensuring smooth continuity during incremental updates.  
\end{itemize}

\section{Related Work}
\label{sec:related_work}

Large-scale environmental mapping is generally based on LiDAR sensors to capture rich spatial information because of their robustness to illumination changes and ability to provide accurate long-range 3D measurements. Existing scene representation and reconstruction methods can be broadly classified into two categories: traditional point cloud fusion techniques and neural-based implicit approaches, while implicit representations can be further divided into dense and sparse formulations.

\subsection{Traditional Point Cloud Fusion}
Conventional fusion methods, such as VDBFusion~\cite{VDBfusion}, VoxBlox~\cite{Voxblox}, and NVBlox~\cite{nvblox}, encode voxelized scene priors in sparse data structures and update TSDF or occupancy volumes via raycasting. These representations are memory-efficient and support fast voxel access, enabling real-time operation with accurate reconstructions when observations are dense. However, in regions with sparse sensor coverage, insufficient ray support leaves voxels under-constrained, leading to gaps, holes, and reduced fidelity.

\subsection{Dense Neural Implicit Mapping}
Recent progress in neural implicit representations was catalyzed by NeRF~\cite{NeRF}, which parameterizes a dense radiance field with a compact Multi-Layer Perceptron (MLP) and achieves high-fidelity rendering of complex scenes. Subsequent works~\cite{mipnerf360, DVGO} focus on improving sampling strategies and representation efficiency, while others~\cite{block-nerf,nerfw} focus on robustness and scalability for large or unbounded scenes. These advances inspired indoor SLAM back-ends to adopt implicit fields~\cite{imap,NeuralRGBD,NiceSLAM}, achieving finer scene representations. However, dense per-ray sampling imposed a high optimization cost, preventing real-time performance even in small indoor scenes.

To reduce this cost, acceleration techniques focus on improving sampling lookup efficiency. Point-NeRF~\cite{pointnerf} and Instant-NGP~\cite{instant-ngp} adopt a more compressed scene representation to accelerate sample retrieval. SLAM methods~\cite{ESLAM, CoSLAM, PointSLAM} quickly incorporated these advances, pushing indoor mapping frame rates towards real-time. For large-scale outdoor LiDAR, LONER~\cite{loner} incorporated NeRF inspired volume rendering into the SLAM back-end. In contrast, Shine-Mapping~\cite{shine} directly supervises the signed distance at sampled 3D points to regress a continuous TSDF field. PIN-SLAM~\cite{pinslam} couples a dense implicit field with a point-based back-end to allow more efficient data retrieval. Subsequent methods~\cite{3dslnr, clidslam, ucpinslam} introduce several variations, while the fundamental back-end training paradigm remains essentially unchanged: they still underutilize explicit geometric priors. Consequently, substantial training effort is spent on learning free space. Dense implicit mapping methods either rely on offline training to obtain the best result or sacrifice mapping fidelity for real-time incremental updates. 

\begin{figure*}[!htbp]
\vspace{0.1cm}
 \centering
    \includegraphics[width=1.0\linewidth]{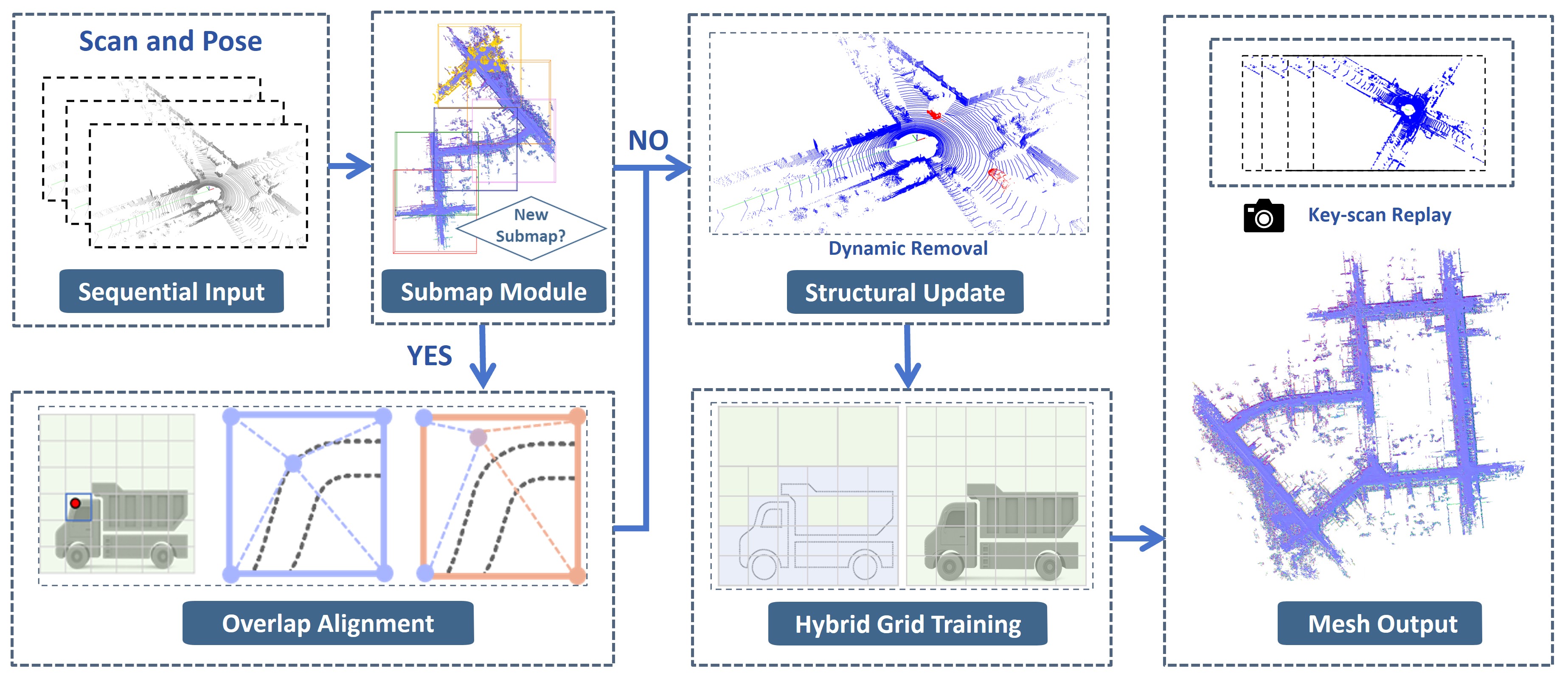}
     \caption{\textbf{Overview of our 3D mapping pipeline.} Sequential LiDAR scans with poses are processed by a submap module that manages submap creation and updates. For scans within an existing submap, the system performs normal mapping with dynamic removal and hybrid grid training. When a new submap is created, an additional overlap alignment module enforces cross-map consistency. Finally, the learned submaps are replayed with key-scans and fused into a global mesh for accurate and efficient large-scale reconstruction.}
    \label{fig:approach-pipeline}
\vspace{-0.2cm}
\end{figure*}

\subsection{Sparse Neural Implicit Mapping}
In contrast to dense implicit representation, sparse neural mapping methods use sparse data structures as priors to guide sampling. Plenoxels~\cite{plenoxels} extends this idea by storing spherical harmonics coefficients in a sparse voxel octree (SVO), achieving better mapping quality and faster convergence speed. Vox-Fusion~\cite{voxfusion} and H2-Mapping~\cite{h2mapping} adopt this idea for indoor reconstruction, sampling only occupied cells to improve both training efficiency and reconstruction quality. NeRF-LOAM~\cite{nerfloam} extends to large-scale outdoor LiDAR by combining an SVO with learnable voxel features to regress a TSDF, yielding higher reconstruction accuracy than dense implicit baselines. However, its efficiency declines with larger scene sizes and the system does not operate in real time. The follow-up work~\cite{s2kan} adopts the same framework, but only replaces the network structure and employs a simple submap design; thus, the method still suffers from periodic runtime oscillations due to frequent submap resets, and extending the framework to real-time large-scale mapping remains a challenge.

\section{METHODOLOGY}


Our pipeline of XGrid-Mapping is illustrated in Fig.~\ref{fig:approach-pipeline}. The system takes sequential LiDAR scans with associated poses as input, after which a submap module determines whether to create a new submap or update the current one (\cref{subsec:submap}). Within each submap, the mapping module first removes dynamic points to filter out moving objects (\cref{subsec:dynamic}), then applies our hybrid grid training strategy (\cref{subsec:training}),  and enforce cross-submap consistency with an overlap alignment module (\cref{subsec:overlapalignment}). 
Finally, the learned submaps are fused into a global mesh with key-scan replay (\cref{subsec:kscan}), yielding accurate large-scale reconstruction results.

\subsection{Submap Module}
\label{subsec:submap}

In each submap $\Omega$ we define our XGrid representation: an explicit sparse grid $\Omega^s$ and an implicit dense grid $\Omega^d$. The explicit sparse grid, implemented with a VDB-based structure~\cite{fVDB}, provides structural priors, while the implicit dense grid, implemented with a multiresolutional hash encoding~\cite{tiny-cuda-nn}, offers a compact and expressive representation of the scene. Each submap has a fixed size $l = (l_x, l_y, l_z)$, which determines the spatial extent of its bounding box. For the first frame, we initialize a submap $\Omega_{0}$. For each subsequent frame, we compute the entry rate $r$ of its point cloud scan, defined as the ratio of points that fall within the current submap. If $r$ falls below the threshold $r_{\mathrm{min}}$, a new submap of the same size $l$ is created. The center $c_{t+1}$ is first defined as the center of this scan. 

When building a new submap $\Omega_{t+1}$, its center $c_{t+1}$ aligns with that of the previous submap $\Omega_{t}$ by accounting for the voxel size $s_v$. Specifically, the nearest integer-multiple offset is determined to align the new submap center with the old one. The aligned center $c^{\prime}_{t+1}$ is then computed as:
\begin{equation}
    c_{t+1}' = c_t + \text{round}\left( \frac{c_{t+1} - c_t}{s_v} \right) \cdot s_v.
\end{equation}
With this alignment, voxel edges are aligned, and the minimum corner of the submap bounding box is defined as:
\begin{equation}
    b_{\min} = c_{t+1}' - \tfrac{1}{2} \, l.
\end{equation}

For real-time efficiency, our sparse grid structure, together with the submap design, effectively overcomes the run-time bottleneck of sparse neural mapping methods\cite{nerfloam, s2kan}.






\subsection{Structural Update with Dynamic Removal}
\label{subsec:dynamic}
Dynamic point removal is crucial for mapping methods built on sparse data structures. VDBFusion~\cite{VDBfusion} and NeRF-LOAM~\cite{nerfloam} introduced a space-curving scheme to suppress dynamic outliers. However, in large outdoor environments, these approaches require stepping along full rays rather than within a truncated narrow band, and they still cannot completely remove dynamic points, making the algorithm less robust. To address this, we adopt the idea from FreeDOM~\cite{FreeDOM}. Starting from scan points as seeds, FreeDOM grows regions by combining raycasting checks with neighborhood expansion, directly labeling scan points as dynamic or static. This design removes the dynamics from the source, making the system more robust.

Formally, given an input scan $S = \{p_i\}$, each point is first transformed into the world frame by pose $T_w$:  
\begin{equation}
    S_w = \{\, T_w \, p_i \mid p_i \in S \,\}.
\end{equation}
The points are then mapped into the local submap coordinate by subtracting the minimum corner of the submap $b_{\min}$:  
\begin{equation}
    S_{\text{loc}} = S_w - b_{\min}.
\end{equation}
The preprocessing step of Dynamic Removal then outputs the static subset:  
\begin{equation}
    S_{\text{static}} \subseteq S_{\text{loc}}.
\end{equation}
Finally, these static points are used to update the sparse grid:  
\begin{equation}
    \Omega^s \leftarrow \text{Update}(\Omega^s, S_{\text{static}}).
\end{equation}

\subsection{Hybrid Grid Training}
\label{subsec:training}
\begin{figure}[t]
 \raggedleft
 \vspace{0.2cm}
 \hspace{-30pt}
    \includegraphics[width=1.0\linewidth]{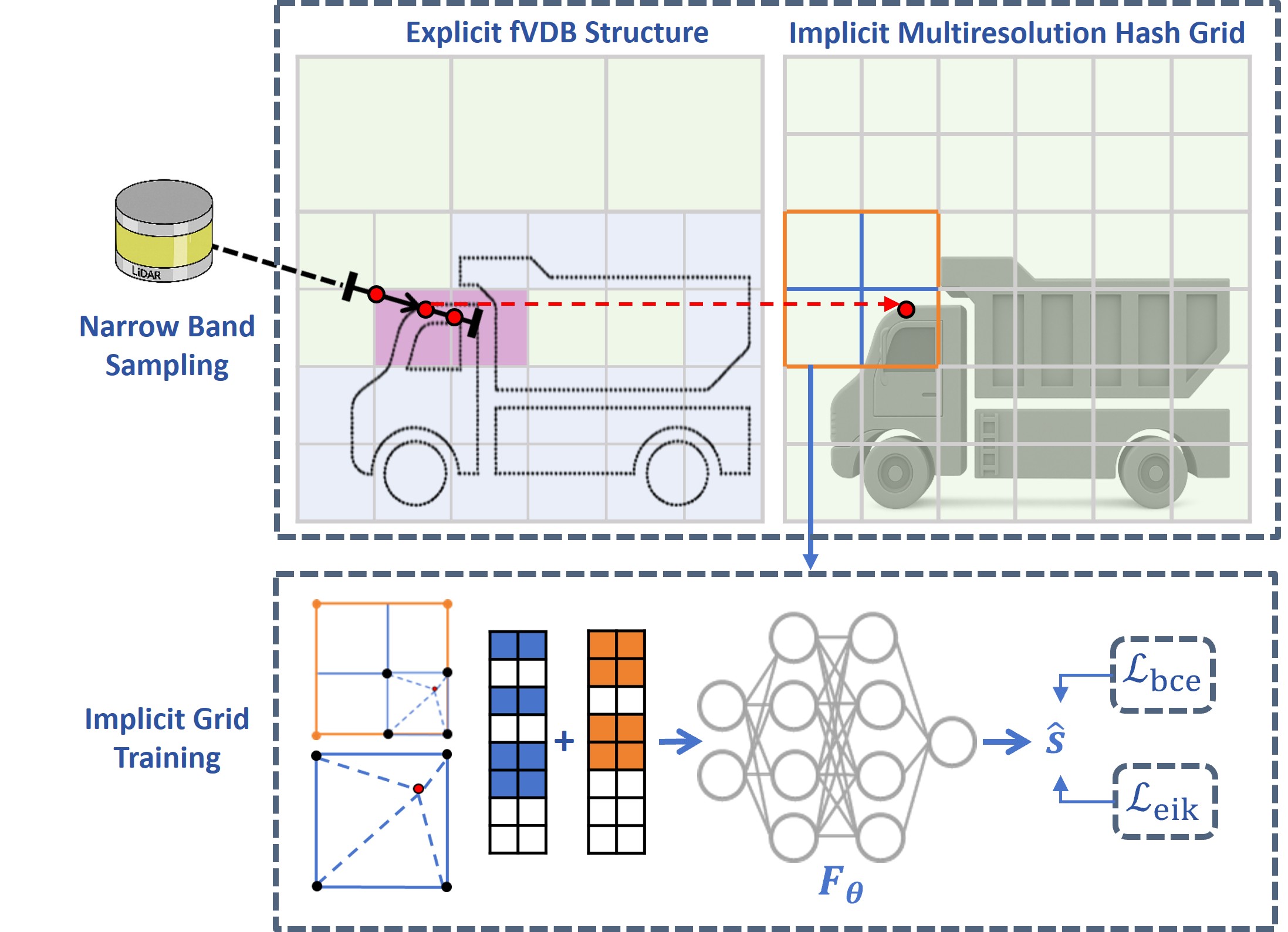}
     \caption{\textbf{Illustration of the hybrid grid training strategy.} Top: narrow-band sampling in the explicit sparse grid, where black arrows indicate rays and red dots denote sampled points, combined with an implicit dense grid for scene representation. Bottom: training performs on voxel vertices (black), where multiscale embeddings are concatenated into vertex features, interpolated to sample points, and passed through an MLP to predict SDF values, optimized with two loss terms.}
    \label{fig:Dual-grid}
\vspace{-0.4cm}
\end{figure}
We demonstrate our training process in Fig.~\ref{fig:Dual-grid}. Our sampling strategy is based on the explicit sparse grid $\Omega^s$. During raycasting, most of the free space is directly constructed by the sparse grid. With the dynamic removal module eliminating moving objects, we can sample more efficiently within the truncated narrow band around the surface. 

Formally, for each LiDAR ray with origin point $o$ and end point $r_i$, the ray can be written as:
\begin{equation}
    r_i(t) = o + t \cdot \frac{r_i}{\lVert r_i \rVert}, \quad t \geq 0,
\end{equation}
where $d_i = \lVert r_i \rVert$ is the measured range and $u_i = \tfrac{r_i}{\lVert r_i \rVert}$ is the unit ray direction. 
Narrow-band samples are generated as:
\begin{equation}
    p_s = o + (d_i + \delta)\,u_i, \quad \delta \sim \mathcal{U}(-T_r,\,T_r),
\end{equation}
where $T_r$ denotes the truncation distance and $\mathcal{U}(\cdot)$ denotes uniform sampling. The ground-truth SDF values for samples are defined as:
\begin{equation}
    s(p_s) = \operatorname{sign}\big(d_i - \lVert p_s - o \rVert\big)\cdot \lVert p_s - r_i \rVert.
\end{equation}

XGrid-Mapping follows a vertex-driven training strategy tailored to the explicit voxel representation. For samples $p_s$, we first locate the voxel cell and retrieve the hash features of its eight corner vertices from the implicit grid. The feature of $p_s$ is then obtained by trilinear interpolation:

\begin{equation}
    f(p_s) = \text{Interp}\big(\{h(v_j)\}_{j=1}^8\big),
\end{equation}
where $h(v_j)$ denotes the hash features of the vertex $v_j$. The interpolated vector is then passed through a shallow shared MLP $F_\theta(\cdot)$ to predict the signed distance value:
\begin{equation}
    \hat{s}(p_s) = F_\theta(f(p_s)).
\end{equation}

For implicit grid training, we first transform the signed distances into occupancy probabilities through a scaled sigmoid function:
\begin{equation}
    o(s) = \frac{1}{1 + \exp\!\big(s / \sigma_t\big)},
\end{equation}
where $\sigma_t$ is a temperature hyperparameter that controls the softness of the truncation. For convenience, we denote $o_s = o(s(p_s))$ and $\hat{o}_s = o(\hat{s}(p_s))$. We then introduce a binary entropy loss by:
\begin{equation}
    \mathcal{L}_{\text{bce}} = -\frac{1}{|\mathcal{P}|}\sum_{p_s \in \mathcal{P}} 
 o_s \log \hat{o}_s + \big(1 - o_s\big) \log \big(1 - \hat{o}_s\big),
\end{equation}
where $\mathcal{P}$ denotes the set of sampled points. In addition, we regularize the predicted SDF with an Eikonal loss imposed directly on the sampled points:
\begin{equation}
    \mathcal{L}_{\text{eik}} = \frac{1}{|\mathcal{P}|}\sum_{p_s \in \mathcal{P}} \Big(\|\nabla \hat{s}(p_s)\|_2 - 1\Big)^2.
\end{equation}
Finally, the overall training loss combines the BCE loss with the Eikonal regularization:
\begin{equation}
\mathcal{L} =
\lambda_{\text{bce}} \, \mathcal{L}_{\text{bce}} 
+ \lambda_{\text{eik}} \, \mathcal{L}_{\text{eik}},
\end{equation}
where $\lambda_{\text{bce}}$ and $\lambda_{\text{eik}}$ are weighting coefficients for the corresponding loss terms.


\subsection{Overlap Alignment}
\label{subsec:overlapalignment}

\begin{figure}[t]
 \centering
    \vspace{0.2cm}
    \includegraphics[width=1.0\linewidth]{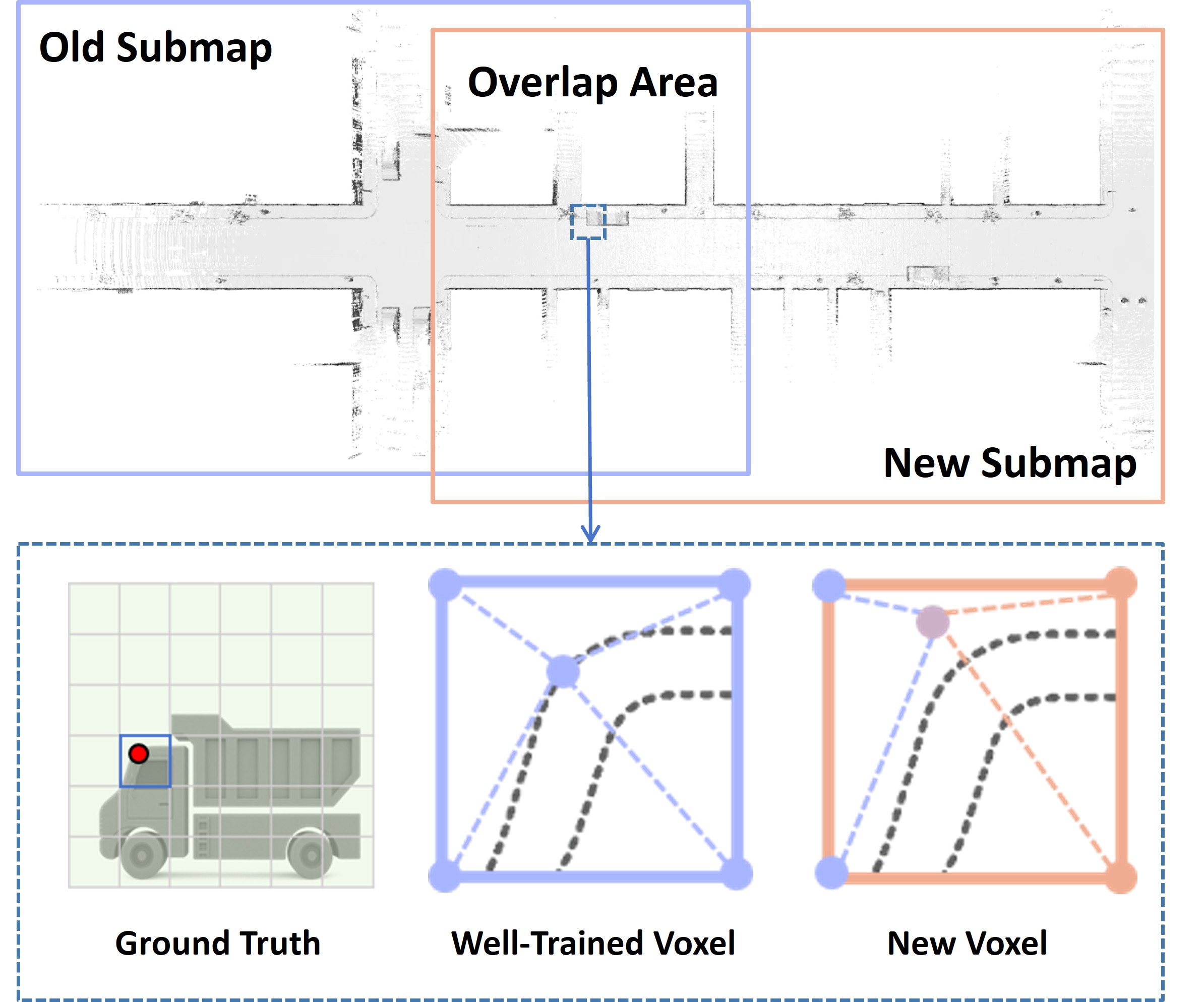}
     \caption{\textbf{Illustration of overlap map distillation strategy.} The upper part illustrates the overlapping region between the old and new submaps, while the lower part shows a point within the overlapping area, where the information of well-trained voxels are used to constrain the learning of voxels in the new submap.}
    \label{fig:distillation strategy}
\vspace{-0.5cm}
\end{figure}

When switching to a new submap, the dense grid is newly initialized and therefore not well trained. This causes slow convergence and prevents the model from fully exploiting observations in the overlapping region with the previous submap. As a result, feature drift and discontinuities may arise, degrading overall reconstruction quality. To address this challenge, we introduce an overlap alignment module, which enforces coherence on voxels shared between consecutive submaps, as illustrated in Fig.~\ref{fig:distillation strategy}. 

Specifically, voxels in the previous sparse grid $\Omega_t^s$ are extracted and inserted into the new grid $\Omega_{t+1}^s$, forming an overlapping set $\mathcal{O}$. For each overlapping voxel $vo \in \mathcal{O}$, we retrieve the well-trained features $h_j^{(t)}$ from the previous grid $\Omega_t^d$ and the newly initialized features $h_j^{(t+1)}$ from the current grid $\Omega_{t+1}^d$ at its eight corner vertices ($j=1,\dots,8$). These features are then aligned by the following loss during overlap frames:  

\begin{equation}
    \mathcal{L}_{\text{align}} = \lambda_{\text{align}} 
    \sum_{vo \in \mathcal{O}} \sum_{j=1}^{8} 
    \left\| h_j^{(t)} - h_j^{(t+1)} \right\|_1 ,
\end{equation}
where $\lambda_{\text{align}}$ is a weighting coefficient.
This loss serves as a distillation objective, encouraging consistent vertex features across submaps and thereby improving continuity and convergence during transitions.

\subsection{Mesh Output with Key-Scan Replay}
\label{subsec:kscan}

For each submap, we maintain a queue of key-scans that are periodically sampled and replayed to provide collective constraints and mitigate catastrophic forgetting. For each frame, we check the distance between the current frame and the previous key-scan. If the distance exceeds a predefined threshold $D_{\mathrm{min}}$, we add it to the key-scan set. When a new submap is created, we perform a replay of the previous submap using the key-scan set. In line with the approach proposed by~\cite{shine}, large-scale scenes often have an excessively large number of key-scans. Therefore, after replaying each submap, we release the key-scan list of that submap to maintain memory efficiency. We then apply Marching Cubes~\cite{marchingcubes} to each submap to extract its surface mesh, which is then merged into the global reconstruction.



\section{EXPERIMENTAL RESULTS AND ANALYSIS}

In this section, we first analyze the efficiency improvements of our algorithm over the sparse implicit baselines, highlighting its advantage in scalability and real-time performance. We then evaluate the mapping quality by comparing our method with open-source state-of-the-art incremental methods, including VDBFusion~\cite{VDBfusion}, SHINE-Mapping~\cite{shine}, NeRF-LOAM~\cite{nerfloam}, and PIN-SLAM~\cite{pinslam}. These baselines are chosen to cover the spectrum from traditional fusion to dense and sparse neural implicit approaches. Performance is assessed across various metrics to demonstrate the advantages of our approach in large-scale reconstruction.
\begin{figure}[!htbp]
 \vspace{0.1cm}
 \centering
    \includegraphics[width=0.98\linewidth]{picture_new/runtime_plot_bold.png}
     \caption{\textbf{Runtime cost of frame mapping over 500 frames.} NeRF-LOAM slows down progressively, while submap-only and FVDB-only incur periodic or gradual overhead. Our design maintains stable real-time performance at $\sim$0.3\,s per frame.}
    \label{fig:runtime}
\vspace{-0.1cm}
\end{figure}

\begin{table}[t]
\vspace{0.1cm}
\centering
\caption{Mapping speed comparison in seconds per frame}
\label{tab:runtime}

\begin{threeparttable}
\renewcommand{\arraystretch}{1.0} 
\setlength{\tabcolsep}{12.5pt}      

\begin{tabular}{@{}ccccc@{}}
\toprule
Methods  & MAI00 & MAI01 & KT03 & KT07 \\
\midrule
Baseline & 4.41 & 1.26 & 4.52 & 6.72 \\
w Submap only & 0.99 & 0.65 & 1.03 & 1.01 \\
w FVDB only & 0.49 & 0.31 & 0.55 & 0.51\\
Ours & \textbf{0.31} & \textbf{0.28} & \textbf{0.29} & \textbf{0.26} \\
\bottomrule
\end{tabular}

\begin{tablenotes}[flushleft]
\footnotesize
\item MAI stands for Maicity. KT stands for KITTI. Lower values indicate faster mapping speed.
\end{tablenotes}
\end{threeparttable}

\vspace{-0.4cm}
\end{table}


\subsection{Experimental Setup}
\subsubsection{Datasets}
To evaluate the performance of XGrid-Mapping, we conducted experiments on three outdoor LiDAR datasets: the Maicity dataset~\cite{Maicity}, the Newer College dataset~\cite{ncd}, and the KITTI odometry dataset~\cite{kitti}. We first adopted the Maicity dataset, a synthetic static benchmark with 64-beam LiDAR scans, which provides a clean reference for controlled evaluation. We then validate the robustness on the more challenging Newer College dataset, collected with handheld LiDAR at Oxford University, which introduces significant measurement noise and slow-moving dynamic objects, such as pedestrians. Finally, to further test scalability and efficiency, we employ the KITTI odometry dataset, the most challenging benchmark with large-scale urban driving sequences featuring long trajectories. Together, these datasets enable a comprehensive evaluation of XGrid-Mapping in static, noisy, and large-scale scenarios.
\begin{figure*}[!htbp]
 \centering
 \vspace{0.2cm}
    \includegraphics[width=1.0\linewidth]{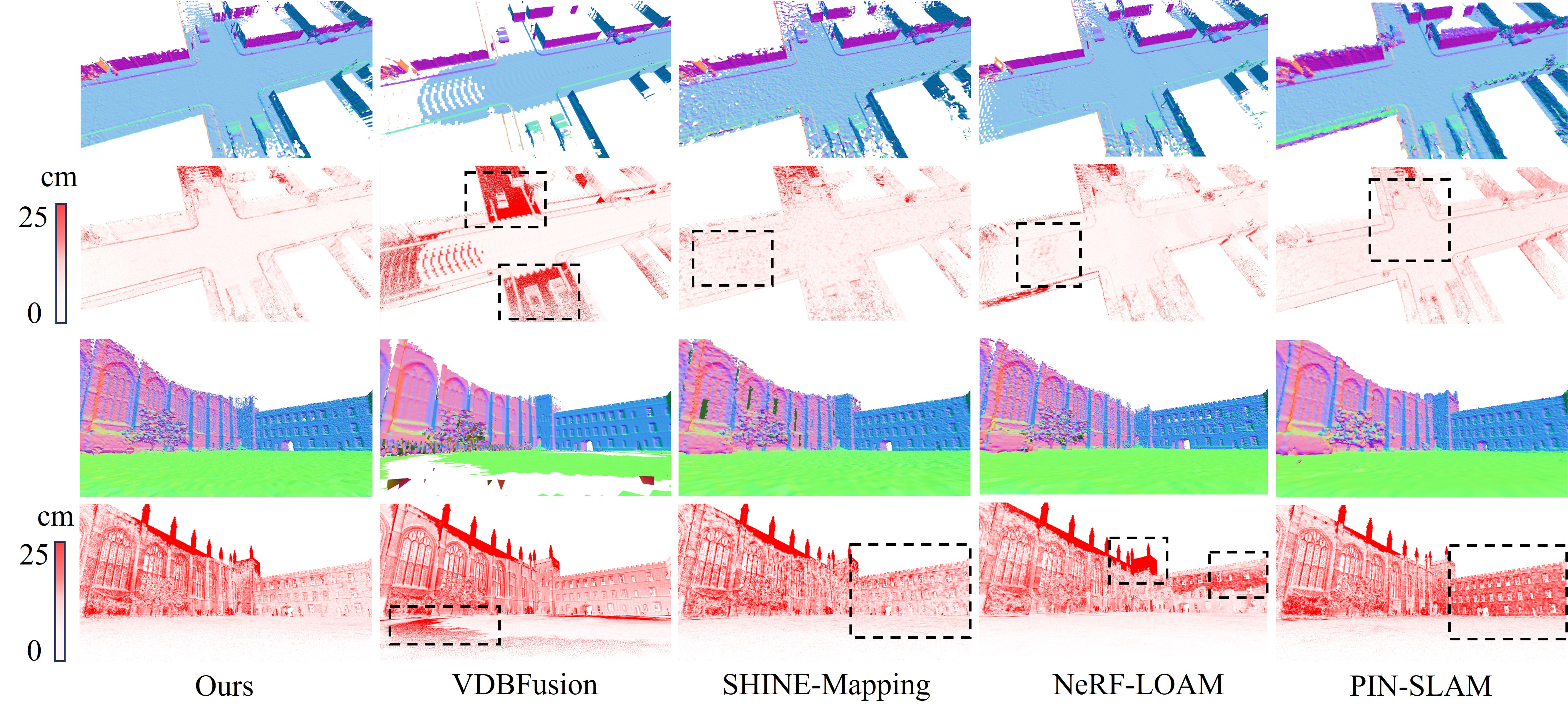}
    \vspace{-20pt}
     \caption{\textbf{Quantitative results.} The first two rows show results on the Maicity dataset~\cite{Maicity}, with surface normals in the first row and error maps in the second row. The last two rows show results on the Newer College dataset~\cite{ncd}, again with surface normals followed by error maps. Redder regions in the error maps indicate larger deviations from the ground-truth mesh, up to 25 cm.}
    \label{fig:eval-m}
\vspace{-0.1cm}
\end{figure*}


\subsubsection{Metrics}
For quantitative evaluation, we adopt widely used reconstruction metrics~\cite{occnet},~\cite{shine},~\cite{nerfloam} to assess the quality of the predicted mesh against the ground truth. Specifically, we evaluate mapping accuracy (Acc., cm), which measures the average distance from predicted points to the ground truth surface, and completeness (Comp., cm), which quantifies how well the predicted reconstruction covers the reference mesh. Additionally, we report the Chamfer-L1 distance (C-L1, cm), a metric that balances accuracy and completeness. Lastly, we include the F-score, which represents the harmonic mean of precision and recall, offering a more balanced assessment of reconstruction quality. 

\begin{table*}[t]
\vspace{0.1cm}
\renewcommand{\arraystretch}{1.3}
\setlength\tabcolsep{8pt}
\centering
\caption{Quantitative reconstruction quality comparison}
\label{tab:mapping}

\begin{threeparttable}
\begin{tabular}{ccccccccc}
\toprule
\multirow{2}{*}[-0.6ex]{{\textbf{Methods}}}
  & \multicolumn{4}{c}{\textbf{Maicity}} 
  & \multicolumn{4}{c}{\textbf{Newer College}} \\
  \cmidrule(lr){2-5}\cmidrule(lr){6-9}
& Acc. \textdownarrow & Comp. \textdownarrow & C-L1 \textdownarrow & F-Score \textuparrow
& Acc. \textdownarrow & Comp. \textdownarrow & C-L1 \textdownarrow & F-Score \textuparrow \\
\midrule
VDBfusion      & 3.44 & 3.65 & 3.55 & 93.3  & \textbf{6.42} & 16.1 & 11.3 & 91.2 \\
SHINE-Mapping  & 4.62 & 2.22 & 3.42 & 92.3  & 7.21 & 11.4 & 9.35 & 91.7 \\
NeRF-LOAM      & 3.60 & 2.07 & 2.84 & 94.2  & 6.92 & 15.7 & 11.3 & 91.8 \\
PIN-SLAM       & 5.60 & 2.70 & 4.15 & 87.6  & 8.84 & 11.7 & 10.3 & 89.7 \\
Ours           & \textbf{3.38} & \textbf{2.01} & \textbf{2.69} & \textbf{95.3}
               & 6.83 & \textbf{10.6} & \textbf{8.72} & \textbf{93.0} \\
\bottomrule
\end{tabular}

\begin{tablenotes}[flushleft]
\footnotesize
\item Metrics are in cm for Acc., Comp., and C\,-L1; F\,-Score is in \%. Maicity uses a 10\,cm F\,-Score threshold; Newer College uses 20\,cm. Best in each column is bolded.
\end{tablenotes}
\end{threeparttable}
\vspace{-0.2cm}
\end{table*}

\subsubsection{Implementation Details}
The entire algorithm is developed in PyTorch, and all experimental results were tested on a desktop system equipped with an NVIDIA RTX 3090. For the multiresolution hash grid, we follow the recommended configuration from tiny-cuda-nn~\cite{tiny-cuda-nn}, using a hash table configuration with 
\( L = 16 \) resolution levels, \( F = 2 \) feature dimensions and \( T = 19 \) total tables. The MLP consists of 2 hidden layers with 256 neurons each. The $voxelsize$ is set to 0.2 meters, and the truncation distance \( T_r \) is set to 0.25 meters. The threshold for submap creation $r_{min}$ is set to 0.75. The submap size is not strictly constrained. To better align the explicit and implicit grids, we set it to match the implicit grid bounding box extent, ensuring that the voxel vertices remain aligned. During training, we apply 5 iterations per frame during incremental mapping, meanwhile in regions with overlapping area, we apply 30 additional iterations to enforce feature consistency.

\subsection{Runtime Analysis}
\label{subsec:timeanalysis}
We benchmark the runtime on the MaiCity and KITTI datasets against the sparse neural-implicit baseline NeRF-LOAM~\cite{nerfloam}. As summarized in Table~\ref{tab:runtime}, our method delivers a clear speedup. ~\cref{fig:runtime} illustrates the trend: NeRF-LOAM starts near 0.5 seconds per frame and drifts to several seconds over time, limiting real-time use. This slowdown arises because its SVO continuously expands with the scene, resulting in expansive ray intersections. A submap-only variant reduces the average runtime but still exhibits periodic spikes right before each submap reset, as accumulated scene growth temporarily increases traversal cost. By contrast, an fVDB-only variant benefits from optimized raycasting that greatly accelerates traversal, leading to a gradual decrease in per-frame cost, yet latency still accumulates as the sequence extends. In contrast, coupling submaps with fVDB maintains a nearly constant per frame cost of around 0.3 seconds, thus sustaining real-time performance and alleviating the runtime bottleneck of sparse neural mapping methods.

\subsection{Mapping Quality}

To ensure a fair comparison, we employ ground-truth poses for all methods and apply an incremental mapping strategy to each method. We follow the unified evaluation protocol adopted in the previous work~\cite{shine, nerfloam, pinslam} to obtain the result. However, to provide a more objective evaluation, we report results on the entire reconstructed scene without applying the intersection-based cropping adopted in previous work~\cite{shine, nerfloam, pinslam}.


 Table~\ref{tab:mapping} reports the quantitative results on the Maicity and Newer College datasets, while Fig.~\ref{fig:eval-m} provides qualitative comparisons through error maps and normal maps. Together, they highlight the distinct behaviors of different approaches. VDBFusion delivers high precision on both datasets; however, its reconstructions are notably incomplete, with holes and missing structures clearly visible on the ground in both datasets. Dense neural implicit mapping methods such as SHINE-Mapping and PIN-SLAM adopt similar sampling strategies and training processes. As shown in the error maps, both produce uneven and irregular ground surfaces, as well as high reconstruction errors on wall regions. Moreover, PIN-SLAM trades reconstruction quality for time efficiency, leading to even worse results. NeRF-LOAM achieves competitive precision by using a sparse neural implicit approach, but its completeness is lower on the Newer College dataset, as its training strategy discards long-range measurements. In contrast, our method consistently delivers the most balanced results, achieving the lowest reconstruction error on planar surfaces while maintaining both high precision and completeness. These findings confirm that our design achieves accurate and complete mapping in static and noisy scenarios.



\subsection{Ablation Study}

For overlap alignment and the key-scan strategy, we use the static Maicity dataset to evaluate their impact under clean conditions. For dynamic removal, we further evaluate on the Newer College dataset, which contains noisy measurements and moving objects.



\begin{figure}[t]
\centering
\begin{minipage}{0.23\textwidth}
    \centering
    \includegraphics[width=\textwidth]{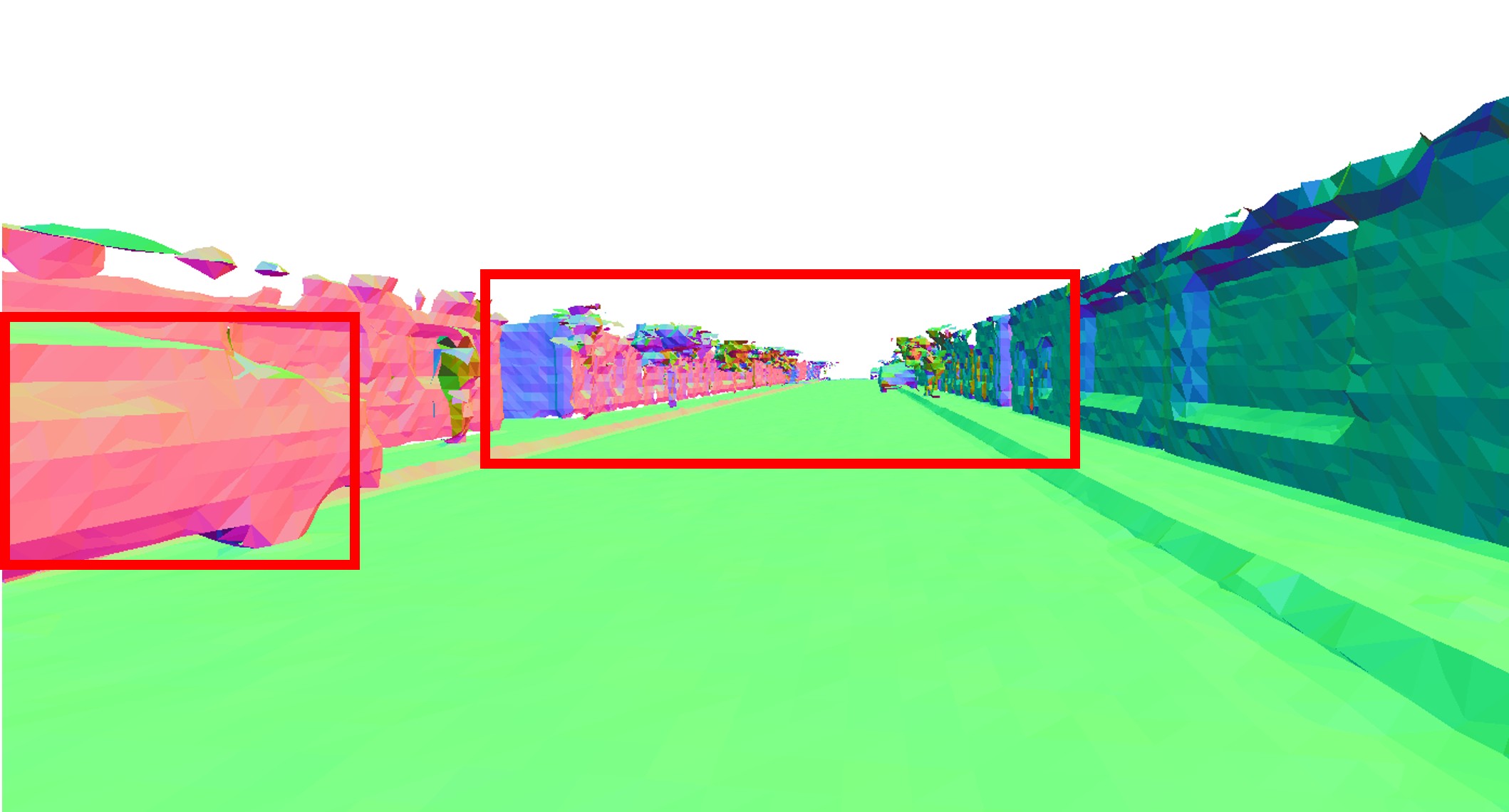}
    \small (a) With alignment
\end{minipage}\hspace{0.01\textwidth}
\begin{minipage}{0.23\textwidth}
    \centering
    \includegraphics[width=\textwidth]{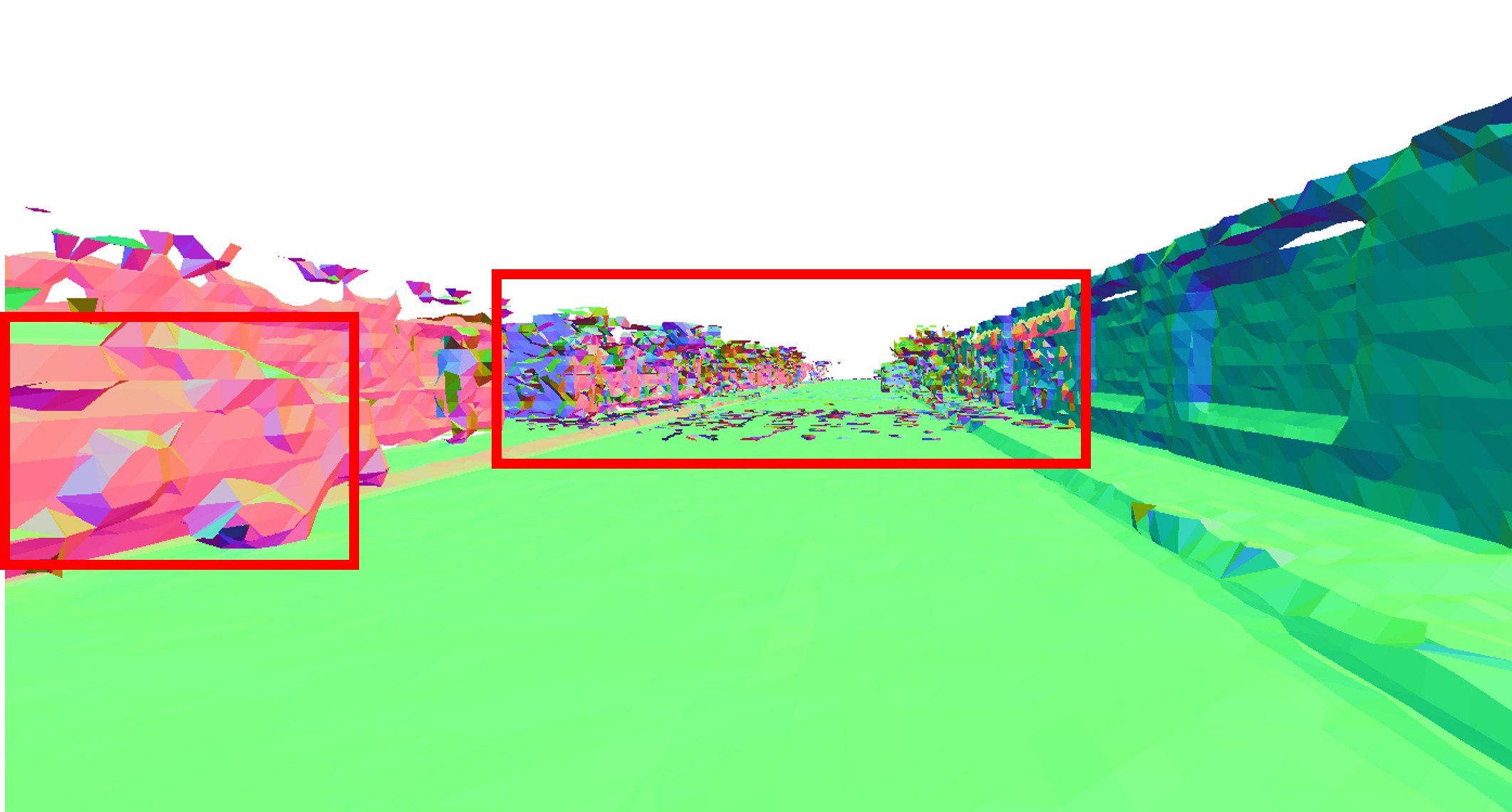}
    \small (b) Without alignment
\end{minipage}


\begin{minipage}{0.23\textwidth}
\vspace{0.3cm}
    \centering
    \includegraphics[width=\textwidth]{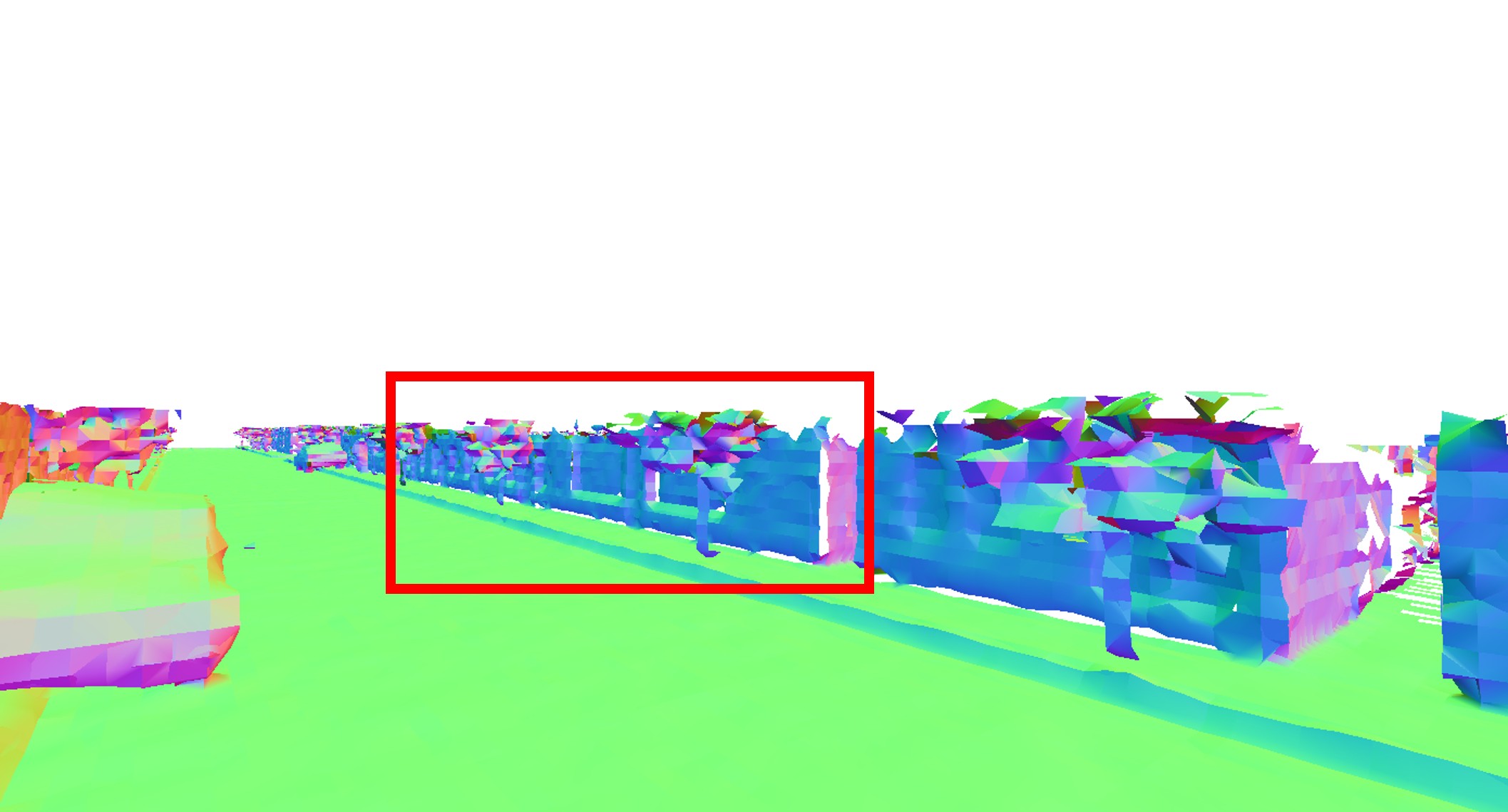}
    \small (c) With key-scan
\end{minipage}\hspace{0.01\textwidth}
\begin{minipage}{0.23\textwidth}
    \centering
    \includegraphics[width=\textwidth]{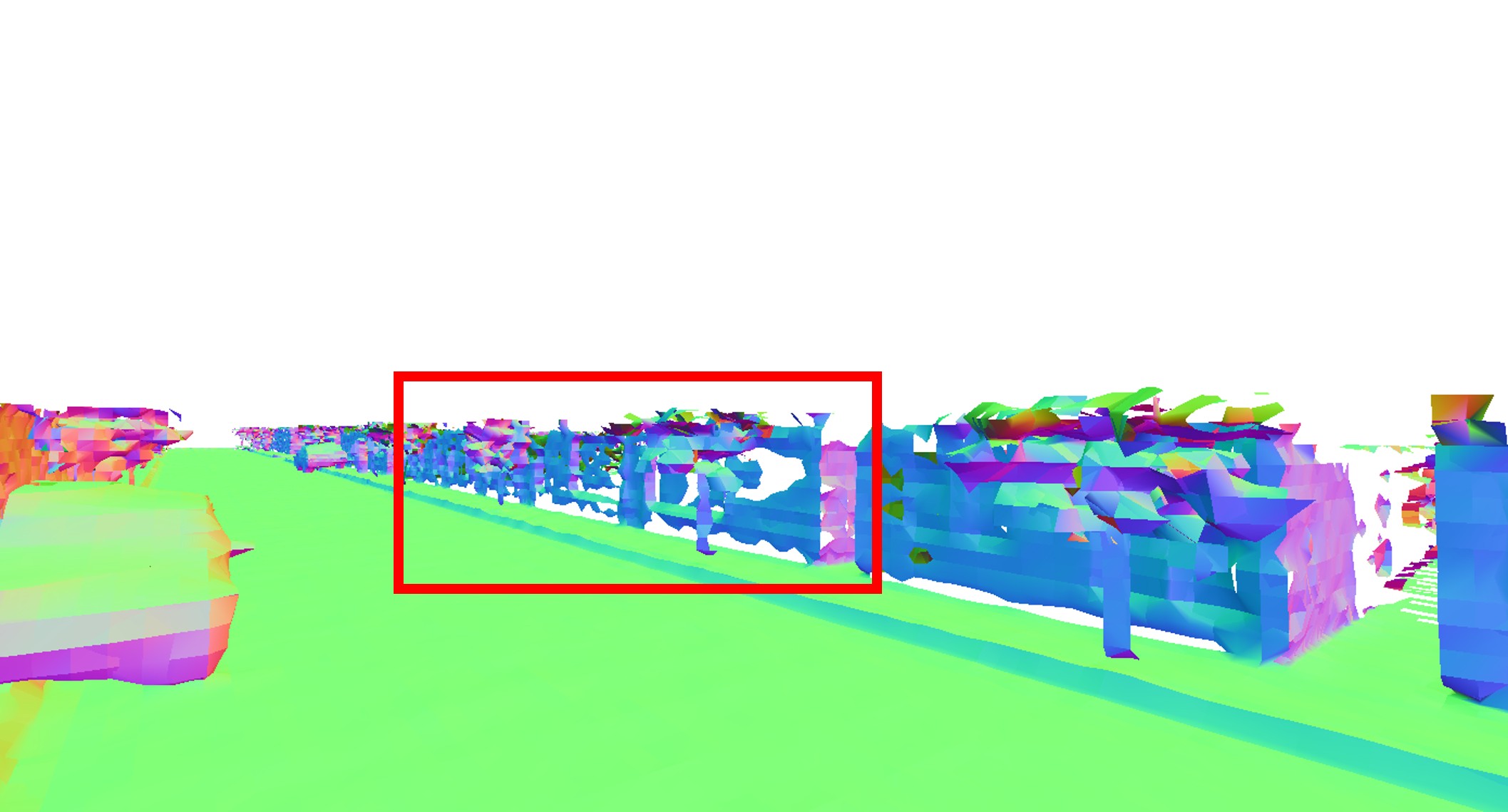}
    \small (d) Without key-scan
\end{minipage}

\caption{\textbf{Effects of overlap alignment and key-scan replay.} 
(a)–(b) show results with/without overlap alignment. 
(c)–(d) show results with/without key-scan replay.}
\label{fig:alignment_keyscan}
\vspace{-0.2cm}
\end{figure}

\begin{table}[t]
\centering
\caption{Ablation study on Maicity 01}
\label{tab:aboverlap}

\begin{threeparttable}
\renewcommand{\arraystretch}{1.2} 
\setlength\tabcolsep{2pt}        

\begin{tabular}{ccccc}
\toprule
Methods & Acc. \textdownarrow & Comp. \textdownarrow & C-L1 \textdownarrow & F-Score \textuparrow \\
\midrule
w/o Both & 3.59 & 3.34 & 3.46 & 94.19 \\
w/o Alignment         & 3.32 & 2.68 & 2.95 & 96.05 \\
w/o Key-Scan          & 3.47 & 3.18 & 3.37 & 94.82 \\
Ours                  & \textbf{2.93} & \textbf{2.62} & \textbf{2.77} & \textbf{96.63} \\
\bottomrule
\end{tabular}

\begin{tablenotes}[flushleft]
\footnotesize
\item The best metric in each column is highlighted in bold. 
\vspace{-0.6cm}
\end{tablenotes}
\end{threeparttable}
\end{table}

\subsubsection{Overlap Alignment}
\label{subsubsec:expoverlap}
Table \ref{tab:aboverlap} presents the quantitative results for the overlap alignment module. Without enforcing alignment, noise accumulates across submaps, convergence remains incomplete, resulting in higher errors in both accuracy and completeness. This observation is further supported by the visualizations in Fig.~\ref{fig:alignment_keyscan}, where misalignment introduces mesh artifacts in the overlapping regions.


\begin{figure}[t]
\vspace{0.2cm}
\centering
\begin{minipage}{0.23\textwidth}
    \centering
    \includegraphics[width=\textwidth]{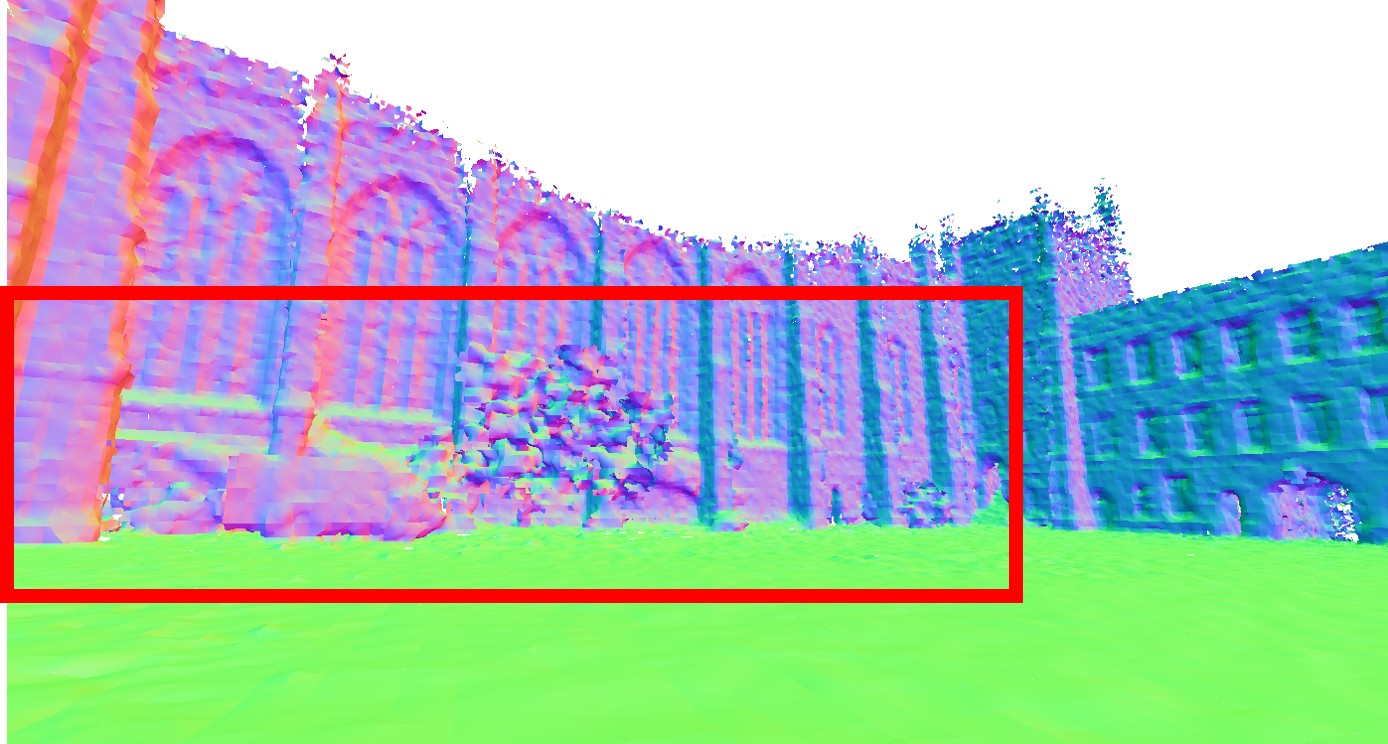}
    \small \text{(a). With dynamic removal}
\end{minipage}%
\hspace{0.01\textwidth} 
\begin{minipage}{0.23\textwidth}
    \centering
    \includegraphics[width=\textwidth]{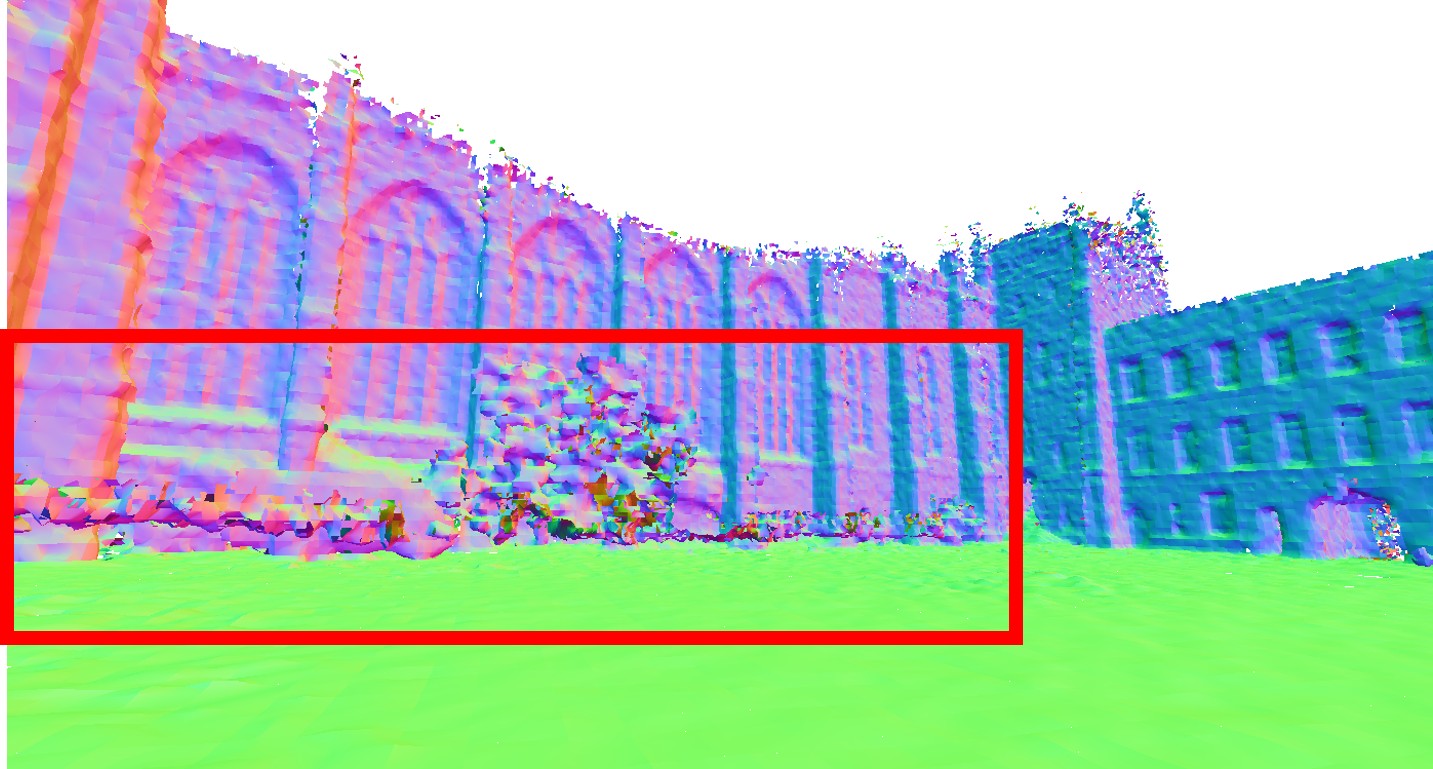}
    \small \text{(b). Without dynamic removal}

\end{minipage}%
\caption{\textbf{Effect of the dynamic removal.} The left image shows our method with dynamic removal, producing clean reconstructions. The right image shows the result without dynamic removal, where walking pedestrians introduce noticeable artifacts and distortions.}
\label{fig:dynamic}
\vspace{-0.1cm}
\end{figure}

\begin{table}[t]
\centering
\caption{Ablation study on Newer College (Cropped)}
\label{tab:ablation_dynamic}

\begin{threeparttable}
\renewcommand{\arraystretch}{1.2} 
\setlength\tabcolsep{2pt}        

\begin{tabular}{ccccc}
\toprule
Methods & Acc. \textdownarrow & Comp. \textdownarrow & C-L1 \textdownarrow & F-Score \textuparrow \\
\midrule
w/o Dynamic Removal & 6.89 & 6.07 & 6.48 & 96.3 \\
w Dynamic Removal   & \textbf{6.70} & \textbf{5.80} & \textbf{6.25} & \textbf{97.3} \\
\bottomrule
\end{tabular}

\begin{tablenotes}[flushleft]
\footnotesize
\item The best metric in each column is highlighted in bold. 
\vspace{-0.4cm}
\end{tablenotes}
\end{threeparttable}
\end{table}

\subsubsection{Key-Scan Strategy}
As shown in Table~\ref{tab:aboverlap}, removing key-scan selection leads to a significant decrease in reconstruction accuracy, with higher Chamfer-L1 errors and lower completeness scores. This indicates that missing key-scans leads to catastrophic forgetting. Fig.~\ref{fig:alignment_keyscan} further illustrates this effect where forgetting issues occur on the wall.

\subsubsection{Dynamic Removal}
Table~\ref{tab:ablation_dynamic} reports the quantitative results on the Newer College dataset. Since the comparison is conducted within our own framework, we crop the ground-truth mesh accordingly for better evaluation. Incorporating our dynamic removal module improves reconstruction quality. The visualizations further highlight its impact: in Fig.~\ref{fig:dynamic}, the reconstruction without dynamic removal exhibits artifacts caused by pedestrians walking through the scene, while our method produces a cleaner mesh without such distortions. 




\section{CONCLUSIONS}


We presented \textbf{XGrid-Mapping}, a submap-based hybrid grid mapping framework. Our design overcomes the computational bottleneck of sparse implicit mapping baseline, making it feasible to combine structural guidance from the sparse grid with the rich expressive power of implicit representations. The distillation-based overlap alignment ensures feature consistency across submaps, while the key-scan replay strategy alleviates catastrophic forgetting during incremental updates. Dynamic removal further stabilizes reconstruction in challenging environments. As a result, our system achieves high mapping quality with improved accuracy and efficiency, providing a reliable foundation for downstream SLAM and robotic perception tasks.

\normalem
\bibliographystyle{IEEEtran}
\bibliography{references}

@article{OctoMap,
author = {Hornung, Armin and Wurm, Kai M. and Bennewitz, Maren and Stachniss, Cyrill and Burgard, Wolfram},
title = {OctoMap: an efficient probabilistic 3D mapping framework based on octrees},
year = {2013},
issue_date = {April     2013},
publisher = {Kluwer Academic Publishers},
address = {USA},
volume = {34},
number = {3},
issn = {0929-5593},
doi = {10.1007/s10514-012-9321-0},
journal = {Auton. Robots},
month = apr,
pages = {189–206},
numpages = {18}
}

@article{VOXELHASH,
author = {Nie\ss{}ner, Matthias and Zollh\"{o}fer, Michael and Izadi, Shahram and Stamminger, Marc},
title = {Real-time 3D reconstruction at scale using voxel hashing},
year = {2013},
issue_date = {November 2013},
publisher = {Association for Computing Machinery},
address = {New York, NY, USA},
volume = {32},
number = {6},
issn = {0730-0301},
doi = {10.1145/2508363.2508374},
journal = {ACM Trans. Graph.},
month = nov,
articleno = {169},
numpages = {11}
}

@article{VDB,
author = {Museth, Ken},
title = {VDB: High-resolution sparse volumes with dynamic topology},
year = {2013},
issue_date = {June 2013},
publisher = {Association for Computing Machinery},
address = {New York, NY, USA},
volume = {32},
number = {3},
issn = {0730-0301},
doi = {10.1145/2487228.2487235},
journal = {ACM Trans. Graph.},
month = jul,
articleno = {27},
numpages = {22}
}

@article{VDBfusion,
author = {Vizzo, Ignacio and Guadagnino, Tiziano and Behley, Jens and Stachniss, Cyrill},
year = {2022},
month = {02},
pages = {1296},
title = {VDBFusion: Flexible and Efficient TSDF Integration of Range Sensor Data},
volume = {22},
journal = {Sensors},
doi = {10.3390/s22031296}
}

@inproceedings{Voxblox,
  author={Oleynikova, Helen and Taylor, Zachary and Fehr, Marius and Siegwart, Roland and  Nieto, Juan},
  booktitle={IEEE/RSJ International Conference on Intelligent Robots and Systems (IROS)},
  title={Voxblox: Incremental 3D Euclidean Signed Distance Fields for On-Board MAV Planning},
  year={2017}
}

@InProceedings{deepsdf,
author = {Park, Jeong Joon and Florence, Peter and Straub, Julian and Newcombe, Richard and Lovegrove, Steven},
title = {DeepSDF: Learning Continuous Signed Distance Functions for Shape Representation},
booktitle = {The IEEE Conference on Computer Vision and Pattern Recognition (CVPR)},
month = jun,
year = {2019}
}

@article{NeRF,
author = {Mildenhall, Ben and Srinivasan, Pratul P. and Tancik, Matthew and Barron, Jonathan T. and Ramamoorthi, Ravi and Ng, Ren},
title = {NeRF: representing scenes as neural radiance fields for view synthesis},
year = {2021},
issue_date = {January 2022},
publisher = {Association for Computing Machinery},
address = {New York, NY, USA},
volume = {65},
number = {1},
issn = {0001-0782},
doi = {10.1145/3503250},
journal = {Commun. ACM},
month = dec,
pages = {99–106},
numpages = {8}
}

@inproceedings{pointnerf,
  title={Point-nerf: Point-based neural radiance fields},
  author={Xu, Qiangeng and Xu, Zexiang and Philip, Julien and Bi, Sai and Shu, Zhixin and Sunkavalli, Kalyan and Neumann, Ulrich},
  booktitle={Proceedings of the IEEE/CVF Conference on Computer Vision and Pattern Recognition},
  pages={5438--5448},
  year={2022}
}

@misc{tiny-cuda-nn,
  author       = {M{\"u}ller, Thomas},
  title        = {tiny-cuda-nn},
  howpublished = {\url{https://github.com/NVlabs/tiny-cuda-nn}},
  year         = {2021},
  month        = apr,
  note         = {Version 2.0. License: BSD-3-Clause},
}

@inproceedings{DVGO,
  author    = {Cheng Sun and Min Sun and Hwann{-}Tzong Chen},
  title     = {Direct Voxel Grid Optimization: Super-fast Convergence for Radiance Fields Reconstruction},
  booktitle = {CVPR},
  year      = {2022},
}

@inproceedings{3dslnr,
  title={3D-SLNR: A Super Lightweight Neural Representation for Large-scale 3D Mapping},
  author={Shi, Chenhui and Tang, Fulin and An, Ning and Wu, Yihong},
  booktitle={Proceedings of the Computer Vision and Pattern Recognition Conference},
  pages={27233--27242},
  year={2025}
}

@article{clidslam,
  title={CLID-SLAM: A Coupled LiDAR-Inertial Neural Implicit Dense SLAM With Region-Specific SDF Estimation},
  author={Jiang, Junlong and Zhang, Xuetao and Sun, Gang and Liu, Yisha and Zhang, Xuebo and Zhuang, Yan},
  journal={IEEE Robotics and Automation Letters},
  year={2025},
  publisher={IEEE}
}

@ARTICLE{ucpinslam,
  author={Gui, Ke and Chen, Zongyang and Chen, Longwen and Wang, Ziyue and Xu, Meilin and Pan, Xianfei},
  journal={IEEE Sensors Journal}, 
  title={UcPIN-SLAM: LiDAR SLAM Using Implicit Neural Representations with Uncertainty Modeling}, 
  year={2025},
  volume={},
  number={},
  pages={1-1},
  doi={10.1109/JSEN.2025.3599584}}

@inproceedings{ESLAM,
  author = {Johari, M. M. and Carta, C. and Fleuret, F.},
  title = {{ESLAM}: Efficient Dense SLAM System Based on Hybrid Representation of Signed Distance Fields},
  booktitle = {Proceedings of the IEEE international conference on Computer Vision and Pattern Recognition (CVPR)},
  year = {2023},
  type = {Highlight}
}

@inproceedings{CoSLAM,
        title={Co-SLAM: Joint Coordinate and Sparse Parametric Encodings for Neural Real-Time SLAM},
        author={Wang, Hengyi and Wang, Jingwen and Agapito, Lourdes},
        booktitle={CVPR},
        year={2023}
}

@inproceedings{PointSLAM,
  author    = {Sandström, Erik and Li, Yue and Van Gool, Luc and R. Oswald, Martin},
  title     = {Point-SLAM: Dense Neural Point Cloud-based SLAM},
  booktitle = {Proceedings of the IEEE/CVF International Conference on Computer Vision (ICCV)},
  year      = {2023}
}

@ARTICLE{loner,
author={Isaacson, Seth and Kung, Pou-Chun and Ramanagopal, Mani and Vasudevan, Ram and Skinner, Katherine A.},
journal={IEEE Robotics and Automation Letters},
title={LONER: LiDAR Only Neural Representations for Real-Time SLAM},
year={2023},
volume={8},
number={12},
pages={8042-8049},
doi={10.1109/LRA.2023.3324521}}

@ARTICLE{h2mapping,
  author={Jiang, Chenxing and Zhang, Hanwen and Liu, Peize and Yu, Zehuan and Cheng, Hui and Zhou, Boyu and Shen, Shaojie},
  journal={IEEE Robotics and Automation Letters}, 
  title={H$_{2}$-Mapping: Real-Time Dense Mapping Using Hierarchical Hybrid Representation}, 
  year={2023},
  volume={8},
  number={10},
  pages={6787-6794},
  doi={10.1109/LRA.2023.3313051}}

@INPROCEEDINGS{block-nerf,
  author={Tancik, Matthew and Casser, Vincent and Yan, Xinchen and Pradhan, Sabeek and Mildenhall, Ben P. and Srinivasan, Pratul and Barron, Jonathan T. and Kretzschmar, Henrik},
  booktitle={2022 IEEE/CVF Conference on Computer Vision and Pattern Recognition (CVPR)}, 
  title={Block-NeRF: Scalable Large Scene Neural View Synthesis}, 
  year={2022},
  volume={},
  number={},
  pages={8238-8248},
  doi={10.1109/CVPR52688.2022.00807}
}

@inproceedings{nerfw,
    author = {Martin-Brualla, Ricardo
                    and Radwan, Noha
             and Sajjadi, Mehdi S. M.
                                    and Barron, Jonathan T.
                   and Dosovitskiy, Alexey
                                    and Duckworth, Daniel},
    title = {{NeRF in the Wild: Neural Radiance Fields for
                                   Unconstrained Photo Collections}},
    booktitle = {CVPR},
    year={2021}
}

@inproceedings{imap,
    title={{iMAP}: Implicit Mapping and Positioning in Real-Time},
    author={Edgar Sucar and Shikun Liu and Joseph Ortiz and Andrew Davison},
    booktitle={Proceedings of the International Conference on Computer Vision ({ICCV})},
    year={2021},
  }

@InProceedings{NeuralRGBD,
    author    = {Azinovi\'c, Dejan and Martin-Brualla, Ricardo and Goldman, Dan B and Nie{\ss}ner, Matthias and Thies, Justus},
    title     = {Neural RGB-D Surface Reconstruction},
    booktitle = {Proceedings of the IEEE/CVF Conference on Computer Vision and Pattern Recognition (CVPR)},
    month     = {June},
    year      = {2022},
    pages     = {6290-6301}
}

@inproceedings{NiceSLAM,
  author    = {Zhu, Zihan and Peng, Songyou and Larsson, Viktor and Xu, Weiwei and Bao, Hujun and Cui, Zhaopeng and Oswald, Martin R. and Pollefeys, Marc},
  title     = {NICE-SLAM: Neural Implicit Scalable Encoding for SLAM},
  booktitle = {Proceedings of the IEEE/CVF Conference on Computer Vision and Pattern Recognition (CVPR)},
  year      = {2022}
}

@inproceedings{voxfusion,
  title={Vox-Fusion: Dense Tracking and Mapping with Voxel-based Neural Implicit Representation},
  author={Yang, Xingrui and Li, Hai and Zhai, Hongjia and Ming, Yuhang and Liu, Yuqian and Zhang, Guofeng},
  booktitle={2022 IEEE International Symposium on Mixed and Augmented Reality (ISMAR)},
  pages={499--507},
  year={2022},
}

@inproceedings{shine,
  title={SHINE-Mapping: Large-Scale 3D Mapping Using Sparse Hierarchical Implicit NEural Representations},
  author={Zhong, Xingguang and Pan, Yue and Behley, Jens and Stachniss, Cyrill},
  booktitle = {Proceedings of the IEEE International Conference on Robotics and Automation (ICRA)},
  year={2023}
}

@inproceedings{nerfloam,
      title={NeRF-LOAM: Neural Implicit Representation for Large-Scale Incremental LiDAR Odometry and Mapping}, 
      author={Junyuan Deng and Qi Wu and Xieyuanli Chen and Songpengcheng Xia and Zhen Sun and Guoqing Liu and Wenxian Yu and Ling Pei},
      booktitle={Proceedings of the IEEE/CVF International Conference on Computer Vision},
      year={2023}
}

@article{pinslam,
  title = {{PIN-SLAM: LiDAR SLAM Using a Point-Based Implicit Neural Representation for Achieving Global Map Consistency}},
  author = {Pan, Yue and Zhong, Xingguang and Wiesmann, Louis and Posewsky, Th{\"o}rbjorn and Behley, Jens and Stachniss, Cyrill},
  journal = {IEEE Transactions on Robotics (TRO)},
  volume = {40},
  pages = {4045--4064},
  year = {2024},
  codeurl = {https://github.com/PRBonn/PIN_SLAM},
  url = {https://www.ipb.uni-bonn.de/wp-content/papercite-data/pdf/pan2024tro.pdf}
}

@article{instant-ngp,
author = {M\"{u}ller, Thomas and Evans, Alex and Schied, Christoph and Keller, Alexander},
title = {Instant neural graphics primitives with a multiresolution hash encoding},
year = {2022},
issue_date = {July 2022},
publisher = {Association for Computing Machinery},
address = {New York, NY, USA},
volume = {41},
number = {4},
issn = {0730-0301},
doi = {10.1145/3528223.3530127},
journal = {ACM Trans. Graph.},
month = jul,
articleno = {102},
numpages = {15}
}

@inproceedings{plenoxels,
title={Plenoxels: Radiance Fields without Neural Networks},
author={Sara Fridovich-Keil and Alex Yu and Matthew Tancik and Qinhong Chen and Benjamin Recht and Angjoo Kanazawa},
year={2022},
booktitle={CVPR},
}

@inproceedings{mipnerf360,
  title={Mip-nerf 360: Unbounded anti-aliased neural radiance fields},
  author={Barron, Jonathan T and Mildenhall, Ben and Verbin, Dor and Srinivasan, Pratul P and Hedman, Peter},
  booktitle={Proceedings of the IEEE/CVF conference on computer vision and pattern recognition},
  pages={5470--5479},
  year={2022}
}

@misc{nvblox,
      title={nvblox: GPU-Accelerated Incremental Signed Distance Field Mapping},
      author={Alexander Millane and Helen Oleynikova and Emilie Wirbel and Remo Steiner and Vikram Ramasamy and David Tingdahl and Roland Siegwart},
      year={2024},
      eprint={2311.00626},
      archivePrefix={arXiv},
      primaryClass={cs.RO}
}

@ARTICLE{s2kan,
  author={Wang, Zhong and Zhang, Lin and Wang, Hesheng},
  journal={IEEE Transactions on Circuits and Systems for Video Technology}, 
  title={S²KAN-SLAM: Elastic Neural LiDAR SLAM With SDF Submaps and Kolmogorov-Arnold Networks}, 
  year={2025},
  volume={35},
  number={8},
  pages={7618-7630},
  doi={10.1109/TCSVT.2025.3550871}}

@ARTICLE{FreeDOM,
  author={Li, Chen and Li, Wanlei and Liu, Wenhao and Shu, Yixiang and Lou, Yunjiang},
  journal={IEEE Robotics and Automation Letters}, 
  title={FreeDOM: Online Dynamic Object Removal Framework for Static Map Construction Based on Conservative Free Space Estimation}, 
  year={2025},
  volume={10},
  number={6},
  pages={5577-5584},
  doi={10.1109/LRA.2025.3560881}}

@article{fVDB,
author = {Williams, Francis and Huang, Jiahui and Swartz, Jonathan and Klar, Gergely and Thakkar, Vijay and Cong, Matthew and Ren, Xuanchi and Li, Ruilong and Fuji-Tsang, Clement and Fidler, Sanja and Sifakis, Eftychios and Museth, Ken},
title = {fVDB: A Deep-Learning Framework for Sparse, Large Scale, and High Performance Spatial Intelligence},
year = {2024},
issue_date = {July 2024},
publisher = {Association for Computing Machinery},
address = {New York, NY, USA},
volume = {43},
number = {4},
issn = {0730-0301},
doi = {10.1145/3658226},
journal = {ACM Trans. Graph.},
month = jul,
articleno = {133},
numpages = {15}
}

@inproceedings{maicity,
author = {I. Vizzo and X. Chen and N. Chebrolu and J. Behley and C. Stachniss},
title = {{Poisson Surface Reconstruction for LiDAR Odometry and Mapping}},
booktitle = {icra},
year = 2021,
url = {http://www.ipb.uni-bonn.de/pdfs/vizzo2021icra.pdf},
codeurl = {https://github.com/PRBonn/puma},
videourl = {https://youtu.be/7yWtYWaO5Nk}
}

@inproceedings{ncd,
author = {Ramezani, Milad and Wang, Yiduo and Camurri, Marco and Wisth, David and Mattamala, Matias and Fallon, Maurice},
title = {The Newer College Dataset: Handheld LiDAR, Inertial and Vision with Ground Truth},
year = {2020},
publisher = {IEEE Press},
doi = {10.1109/IROS45743.2020.9340849},
booktitle = {2020 IEEE/RSJ International Conference on Intelligent Robots and Systems (IROS)},
pages = {4353–4360},
numpages = {8},
location = {Las Vegas, NV, USA}
}

@article{kitti,
  title={Are we ready for autonomous driving? The KITTI vision benchmark suite},
  author={Andreas Geiger and Philip Lenz and Raquel Urtasun},
  journal={2012 IEEE Conference on Computer Vision and Pattern Recognition},
  year={2012},
  pages={3354-3361},
  url={https://api.semanticscholar.org/CorpusID:6724907}
}

@InProceedings{occnet,
author = {Mescheder, Lars and Oechsle, Michael and Niemeyer, Michael and Nowozin, Sebastian and Geiger, Andreas},
title = {Occupancy Networks: Learning 3D Reconstruction in Function Space},
booktitle = {Proceedings of the IEEE/CVF Conference on Computer Vision and Pattern Recognition (CVPR)},
month = {June},
year = {2019}
}

@article{marchingcubes,
author = {Lorensen, William E. and Cline, Harvey E.},
title = {Marching cubes: A high resolution 3D surface construction algorithm},
year = {1987},
issue_date = {July 1987},
publisher = {Association for Computing Machinery},
address = {New York, NY, USA},
volume = {21},
number = {4},
issn = {0097-8930},
url = {https://doi.org/10.1145/37402.37422},
doi = {10.1145/37402.37422},
journal = {SIGGRAPH Comput. Graph.},
month = aug,
pages = {163–169},
numpages = {7}
}
\end{document}